\documentclass[10pt,twocolumn,letterpaper]{article}

\usepackage{cvpr}
\usepackage{times}
\usepackage{epsfig}
\usepackage{graphicx}
\usepackage{amsmath}
\usepackage{amssymb}
\usepackage{mathtools}
\usepackage{colortbl}
\usepackage{tabularx}
\usepackage[table]{xcolor}

\usepackage[pagebackref=true,breaklinks=true,letterpaper=true,colorlinks,bookmarks=false]{hyperref}

 \cvprfinalcopy 

\def\cvprPaperID{685} 

\ifcvprfinal\fi
\begin{document}

\title{Action Unit Detection with Region Adaptation, Multi-labeling Learning and Optimal Temporal Fusing }

\author{
Wei Li\\
\normalsize Dept of Electrical Engineering\\
\normalsize CUNY City College\\
\normalsize New York, USA\\
{\tt\small wli3@ccny.cuny.edu}
\and
Farnaz Abtahi\\
\normalsize Dept of Computer Science\\
\normalsize CUNY Graduate Center\\
\normalsize New York, USA\\
{\tt\small fabtahi@gradcenter.cuny.edu}
\and
Zhigang Zhu\\
\normalsize Dept of Computer Science\\
\normalsize CUNY Graduate Center and City College\\
\normalsize New York, USA\\
{\tt\small zhu@cs.ccny.cuny.edu}
}

\maketitle

\begin{abstract}
Action Unit (AU) detection becomes essential for facial analysis. Many proposed approaches face challenging problems in dealing with the alignments of different face regions, in the effective fusion of temporal information, and in training a model for multiple  AU labels. To better address these problems, we propose a deep learning framework for AU detection with region of interest (ROI) adaptation, integrated multi-label learning, and optimal LSTM-based temporal fusing.  First, ROI cropping nets (ROI Nets) are designed to make sure specifically interested regions of faces are learned independently; each sub-region has a local convolutional neural network (CNN) - an ROI Net, whose convolutional filters will only be trained for the corresponding region. Second, multi-label learning is employed to integrate the outputs of those individual ROI cropping nets, which learns the inter-relationships of various AUs and acquires global features across sub-regions for AU detection. Finally, the optimal selection of multiple LSTM layers to form the best LSTM Net is carried out to best fuse temporal features, in order to make the AU prediction the most accurate. The proposed approach is evaluated on two popular AU detection datasets, BP4D and DISFA, outperforming the state of the art significantly, with an average improvement of around 13\% on BP4D and 25\% on DISFA, respectively. 

\end{abstract}

\section{Introduction}

Action Units (AUs) are the basic facial movements that work as the building blocks in formularizing multiple facial expressions. The successful detection of AUs will greatly facilitate the analysis of the complicated facial actions or expressions. AU detection has been studied for decades as one of the basic facial computing problems and many interesting approaches have been proposed. Classical approaches in AU detection either focus on facial landmark-based local features or appearance-based global features. A number of deep learning approaches have also been proposed to learn deeper facial representations that result in better AU detection.

However, some essential problems are still not solved completely. Due to different features for different facial components, individual AUs may need to be considered separately. One image may include multiple AUs, therefore whether training single AU or multi-label AUs has to be analyzed. Since all actions appear in a temporal instead of just static mode, fusing temporal information becomes necessary. So, to achieve the best AU detection performance,  all the three aspects need to be considered.

Since CNNs have proved to be a powerful tool in solving many image-based tasks and several novel deep structures and frameworks have been proposed, we choose these deep learning models to tackle the AU detection problems. Recently, region-based processing is used in the fast/faster RCNN for prediction of object's bounding box or objectiveness probability in \cite{p2,p7}. This inspired us to design separate networks to learn features for different regions of interest. The success in applying LSTM (long and short term memory) in image caption generation \cite{p38} and human action recognition \cite{p33,p36} led us to believe that it is a good temporal information fusing kernel which may be also useful for facial AU detection. 

 \begin{figure*}[thpb]
      \centering
      \includegraphics[scale=.45]{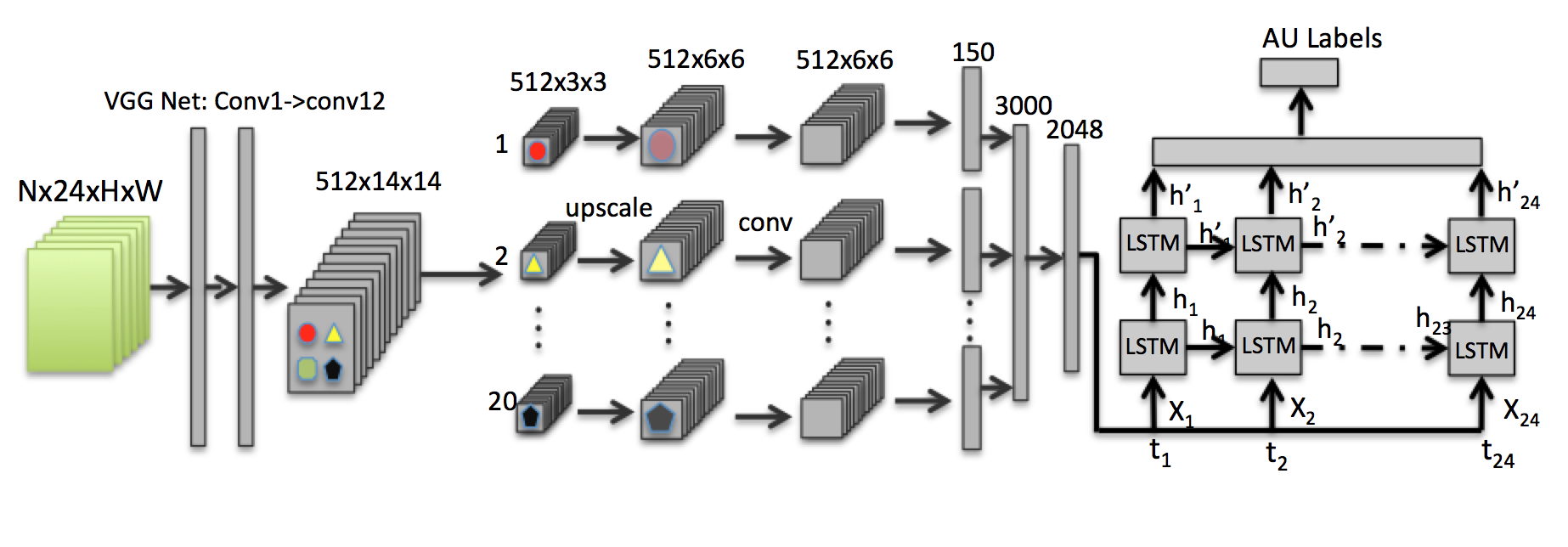}
      \caption{Framework of the proposed neural network with VGG Net, ROI Nets and LSTM Net}
      \label{fig_cnn_all}
   \end{figure*}

After identifying the three problems and being inspired by these RCNN and LSTM approaches, we designed an adaptive region cropping based multi-label learning deep recurrent net. The structure of the proposed neural network is shown in Figure \ref{fig_cnn_all}. There are some unique features of the proposed network. Unlike conventional CNNs where the same convolutional filters are shared within the same convolutional layers, we crop individual regions of interest(ROIs) based on facial landmarks from all the feature maps. The red circle, for instance, represents an area of interest. So, these ROIs are learned individually and therefore important areas will be able to receive special attention. To fuse the temporal information of expressions, the features from the final fully connected layer are fed to several stacks (two in the figure for illustration purpose only) of LSTM layers. Then, the temporal features are used to predict all AUs simultaneously. Through this structure, our network can handle both the adaptive region learning and the temporal fusing problems. 

Comparing to existing approaches, our approach has the following unique contributions:

1) A set of adaptive ROI cropping nets (ROI Nets) is designed to learn regional features separately. In the proposed network, each ROI has a local convolutional neural network. The convolutional filters will only be trained for corresponding regions.

2)Multi-label learning is employed to integrate the outputs of those individual ROI cropping nets, which learns the inter-relationships of various AUs and acquires global features across sub-regions for AU detection.  Multi-label and single AU based methods are compared. With additional AU correlations and richer global features, the multi-label learning approach shows slightly better performance.

3) An LSTM-based temporal fusion recurrent net (LSTM Net) is proposed to fuse static CNN features, which makes the AU predictions more accurate than with static images only. 

This paper is organized as the follows. In Section 1, we have introduced the problems in AU detection and the basic idea of our proposed approach. In Section 2, we review the related work on AU detection including both traditional and deep learning approaches. We then explain our proposed region learning based CNN network in Section 3. Section 4 describes the way the temporal information of the CNN features is fused with the LSTM layers. Experimental results are included in Section 5 where we evaluate our proposed approach in terms of regions cropping, multi-label learning and temporal fusion, and performance comparison against baseline approaches are also given. We conclude the paper in Section 6.

\section{Related Work}

AU detection has been studied for decades and several approaches have been proposed for this problem. Facial key points (landmark points) play an important role in AU detection. Two types of features were usually used in landmark-based approaches. Landmark geometry features were obtained by measuring the normalized facial landmark distances and the angles of the Delaunay mask formed by the landmark points. On the other hand, landmark texture features were obtained by applying multiple orientation Gabor filters to the original images. Many conventional approaches \cite{p10, p11, p14,  p16, p17, p18, p24, p30} were designed by employing texture features near the facial key points. Valstar et al. \cite{p9} analyzed Gabor wavelet features near 20 facial landmark points. The features were then selected and classified by Adaboost and SVM classifiers. Since landmark geometry has been found robust in many AU detection methods, Fabian et al. \cite{p12} et al proposed an approach for fusing the geometry and local texture information.  Zhao et al.\cite{p13} proposed the Joint Patch and Multi-label Learning (JPML) method for AU detection. Similarly, landmark-based regions were selected and SIFT features were used to represent the local patch. Overall, the conventional approaches focused on designing artificial features near facial areas of interest. The appearance changes, representing the motion of the landmark points, give an indication of the facial action units. In addition to facial AU detection, some researchers have also focused on other related problems. Song et al. \cite{p23} investigated the sparsity and co-occurrence of action units. Wu et al.\cite{p19} explored the joint of action unit detection and facial landmark localization and showed that the constraints can improve both AU and landmark detection. Girard et al.\cite{p20} analyzed the effect of different sizes of training datasets on appearance and shape-based AU detection. Gehrig et al.\cite{p21} tried to estimate action unit intensities by employing linear partial least squares to regress intensities in AU related regions.     

Over the last few years, we have witnessed that CNNs boost the performance in many computer vision tasks. Compared to most conventional artificially designed features, CNNs can learn and reveal deeper information from training images. Deep learning has also been employed for AU detection \cite{li2017eac}. Two pieces of the most recent work on the use of deep learning for AU detection are noteworthy. Zhao et al.\cite{p22} used a deep learning approach by dividing the aligned face images into 8x8 blocks. These 64 separate areas are then learned separately. However, although this approach works well for each individual part of a face, it highly relied on face alignment. Additionally, treating all blocks equally may degrade the importance of some regions. Chu et al. \cite{p31} proposed a hybrid approach for combining CNN and LSTM to learn a better representation of an AU sequence. Due to the fusion of both spatial CNN and temporal features, the AU detection performance in this work has improved significantly compared to existing approaches. However, the proposed network is a conventional CNN, which is unable to extract local features from specific regions. Jaiswal et al \cite{jaiswal2016deep} proposed a dynamic appearance and shape based deep learning approach. A shallow region and shape mask CNN is employed to learn the static feature while LSTM is used to extract a dynamic feature from the trained CNN model. In our work, we have designed a CNN which can not only focus on different facial regions independently but also fused the temporal features using recurrent networks.    

 \section{Region of Interest Learning: ROI Nets}
CNNs have recently been the most popular tool for image understanding. In a classic CNN structure, a convolutional layer is composed of multiple filters and activation functions. The convolutional filters cover the entire image and generate the feature map. In this manner, convolutional filters are shared by all the regions of the feature maps. This approach is effective in dealing with general image feature detection, but for some tasks in which individual local regions should be treated differently, sharing the same set of filters for the entire image is not an effective approach.  As most traditional approaches tried to find local SIFT or Gabor features near facial landmark points, we would like to learn local CNN features in these regions of interest (ROIs). 

\begin{figure}[thpb]
      \centering
      \includegraphics[scale=.28]{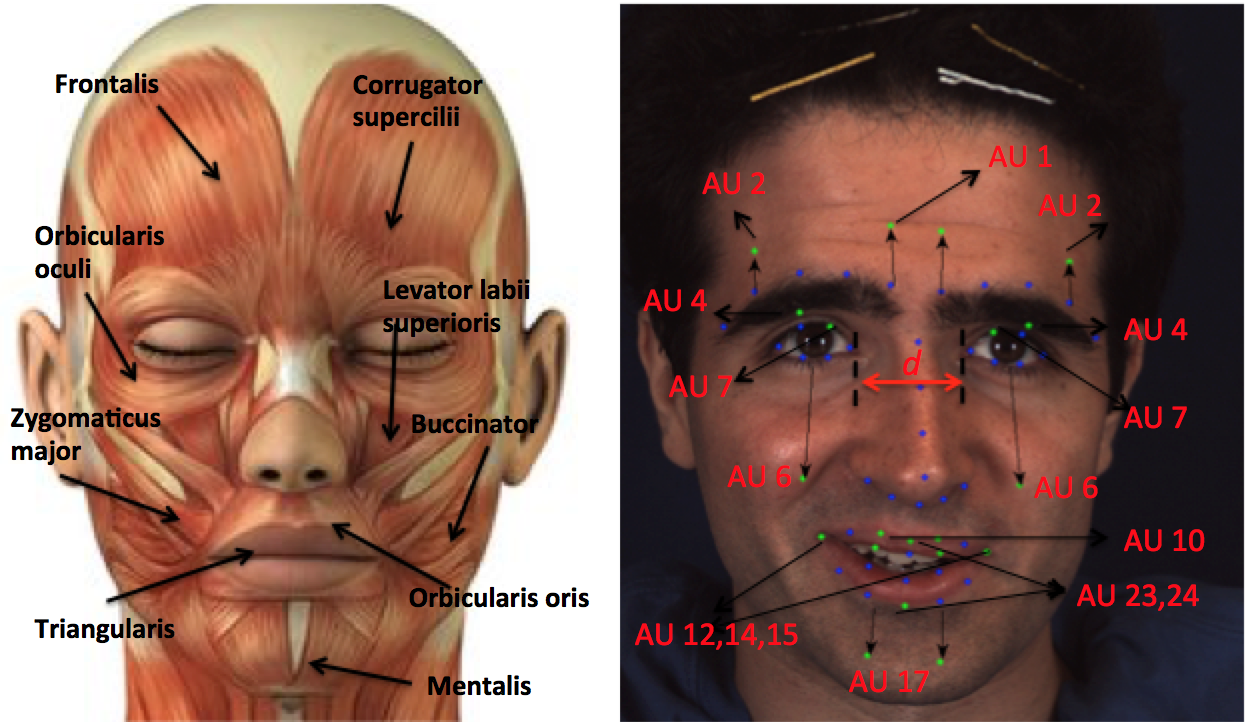}
      \caption{ROI center selection based on muscles and landmarks on one BP4D}
      \label{fig_au_face}
   \end{figure}

We use the BP4D dataset for AU detection which includes 12 AUs. The index, name and corresponding muscles of each AU are illustrated in Table \ref{tab_au_define}  for all the 12 AUs. The corresponding 2D positions of these AUs are shown in Figure \ref{fig_au_face}. We first use a landmark detection algorithm \cite{p25} to find the facial landmark points (blue points in Figure \ref{fig_au_face} right). We choose the AU centers based on the positions of the related muscles (Figure \ref{fig_au_face} left), which are adjusted from face to face using the detected facial landmark points.  Note that some landmark points are not in the centers of facial action muscle regions but they are close to them and can be used to locate the muscles. In the end, the center of an AU is either at a landmark point or a certain distance away from a landmark point, as shown a pair of blue-to-green point in the figure, and we used 20 landmark points in total. 


\begin{table}
\caption{Rules for defining AU centers}
\label{tab_au_define}
\begin{center}
\tabcolsep=0.10cm
\rowcolors{2}{gray!25}{white}

\begin{tabular}{|c|c|c|}
\hline
AU index & Au Name  & Muscle name \\
\hline
1 & Inner Brow Raiser & Frontalis\\
2 & Outer Brow Raiser & Frontalis \\
4 & Brow Lowerer & Corrugator supercilii\\
6 & Cheek Raiser & Orbicularis oculi\\
7 & Lid Tightener & Orbicularis oculi\\
10 & Upper Lip Raiser & Levator labii superioris\\
12 & Lip Corner Puller & Zygomaticus major \\
14 & Dimpler & Buccinator\\
15 & Lip Corner Depressor & Triangularis\\
17 & Chin Raiser & Mentalis\\
23 & Lip Tightener & Orbicularis oris\\
24 & Lip Pressor & Orbicularis oris\\
\hline
\end{tabular}
\end{center}
\end{table}

Knowing the landmark positions, we can then design the neural network cropping layers to form the ROI Nets. We use VGG \cite{p8} as the base for our ROI Net due to its simple structure and excellent performance in object classification. We also choose the 12th convolutional layer as the feature map for cropping. We finally crop the face into 20 ROIs for separate AU learning. In other words, the 20 green points in Figure \ref{fig_au_face} are regarded as ROI AU centers. 

The corresponding positions of AU centers in the feature map can be found based on the ratio of the original image size (224$\times$224) and the feature map size (14$\times$14). Based on the 512$\times$14$\times$14 feature maps as well as the 20 AU centers, we take a total of 20 sub-regions (centered at the selected AU landmark centers), each of  512$\times$3$\times$3, as the input for cropping layers to form the ROI-Nets, 20 in total. For each individual region learning network, the input size of 3$\times$3 might not be able to represent the region well. So, upsampling layers are added to upscale to 6$\times$6 before the convolutional layers. The final adaptive region learning structure is shown in the middle part of Figure \ref{fig_cnn_all}. After local learning with the ROI Nets, we use a fully connected feature vector to represent the local regional features. Then we can either pair the symmetrical features for single AU detection or concatenate all the fully connected features for multi-label AU detection. We will conduct a further comparison on this selection in Section 4. 

 By designing the ROI Nets, we can train separate filters for the AUs. This may make the feature learning adaptive to different local facial properties. Comparison of the ROI learning and conventional CNN learning will be performed in the evaluation section (Section 4).

\section{Temporal Fusing: LSTM Net}

A facial action always has a temporal component when using a video sequence as the input, hence knowing the previous states of a facial expression can definitely improve the AU detection. However, one of the limitations of the CNN structure is the lack memory of previous states.  Regular CNNs are only able to process a single image at a time. To deal with a sequence of images, C3D \cite{p32}, which is basically a 3D version of CNN, has been proposed. C3D can deal with sequential images \color{black} but the number of input images is fixed. \color{black}The training of a C3D is very time-to consume too. Another huge shortage of C3D is, compared to using regular CNN, the lack of existing pretrained models similar to VGG \cite{ p8}, GoogleLeNet \cite{p35} and ResNet \cite{p26}, which can all provide very good initial parameters as a starting point for training. The current best network for temporal fusion is the \color{black}Long Short Term Memory (LSTM) network \cite{LSTM_paper}.\color{black}   As a recurrent net, it can memorize the previous features and states, which can help current feature learning and estimation. It also has gate structures to make it suitable for long time and short time temporal feature learning. LSTM has also proved to be effective in action recognition \cite{p36}. 

 \begin{figure}[thpb]
      \centering
      \includegraphics[scale=.375]{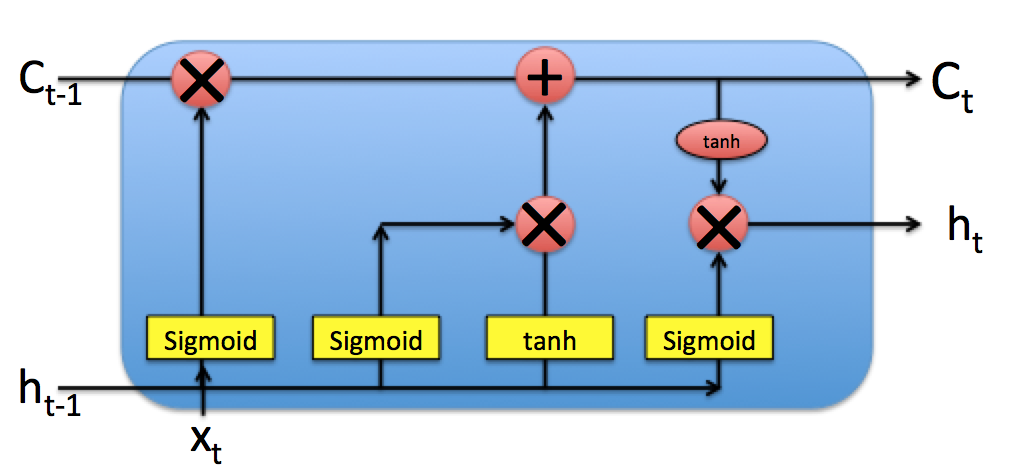}
      \caption{Structure of a simple LSTM block.}
      \label{lstm_struct}
   \end{figure}

The structure of an LSTM block is shown in Figure \ref{lstm_struct}. In the LSTM block, $C_{t-1}$  and $C_t$ are the cell state parameters at the previous and the current times, the long and short memories are described by the \color{black}cell state vector $C_t$\color{black}. The cell states store the memory parameters in LSTM. At each time step, an LSTM kernel will take the previous output $h_{t-1}$ and the new input $x_{t}$ to generate the new output $h_{t}$ through gates, which is shown as yellows blocks in figure \ref{lstm_struct}. Meanwhile, the cell state gets updated. \color{black}A new input feature \color{black}fed to a LSTM block will go through three steps. First, the LSTM has to decide what information to obtain/forget from the old cell state. This is based on the previous LSTM output $h_{t-1}$ and new input feature $x_{t}$. The forget vector $f_{t}$ follows equation \ref{lstm_cellstate}: 

\begin{equation}
\label{lstm_cellstate}
f_{t}=\sigma (W_{f} \cdot[h_{t-1},x_{t}]+b_{f} )
\end{equation}

\noindent where $W_{f}$ and $b_{f}$ are the forget gate parameters. \color{black}The next step is to update the cell state for future use. The new cell state $C_{t}$ is determined by two elements: previous partially saved cell state $C_{t-1}$, current LSTM input $x_{t}$ and previous output $h_{t-1}$. The last two vectors need to go through an ``input gate" and a $tanh$ activation function. The updated cell state can be obtained using equation \ref{lstm_updatecell3}:

\begin{equation}
\label{lstm_updatecell3}
C_{t}=f_{t}\ast C_{t-1}+i_{t}\ast\check{C_{t}}
\end{equation}
\noindent where $i_{t}$ is the merged input of $x_{t}$ and $h_{t-1}$ defined by equation \ref{lstm_updatecell1},
\begin{equation}
\label{lstm_updatecell1}
i_{t}=\sigma (W_{i} \cdot[h_{t-1},x_{t}]+b_{i} )
\end{equation}
\noindent where $W_{i}$ and $b_{i}$ are the input gate parameters.
\noindent $\check{C_{t}}$ in equation \ref{lstm_updatecell3} is the candidate cell state for generating final cell state and output which we can regard as a temporal cell state parameter, following equation \ref{lstm_updatecell2}: 
\begin{equation}
\label{lstm_updatecell2}
\check{C_{t}}=tanh(W_{c} \cdot[h_{t-1},x_{t}]+b_{c} )
\end{equation}
\noindent where $W_{c}$ and $b_{c}$ are the candidate gate parameter.

Finally, we  generate the current output $h_{t}$ for the LSTM based on the updated cell state $C_{t}$, the current input feature $x_{t}$ and the previous output $h_{t-1}$, which can be described by equation \ref{lstm_output}:

\begin{equation}
\label{lstm_output}
h_{t}=\sigma (W_{o} \cdot[h_{t-1},x_{t}]+b_{o} )\cdot tanh(C_{t})
\end{equation}
\noindent where $W_{o}$ and $b_{o}$ are the output gate parameters. \color{black}  Meanwhile, the output $h_{t}$ and the cell $C_{t}$ are passed to next time output generation.

\begin{figure}[thpb]
      \centering
      \includegraphics[scale=.45]{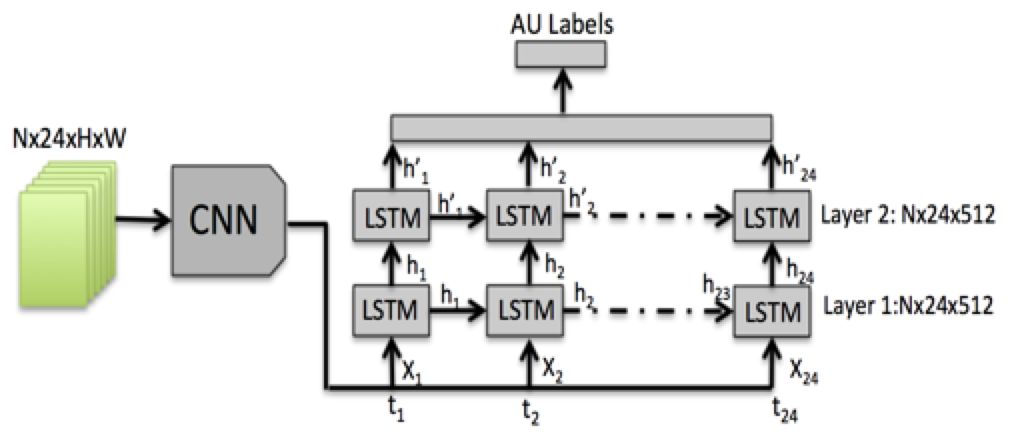}
      \caption{Connection of CNN and LSTM (24 means number of frames in a sequence)}
      \label{lstm_array}
\end{figure}
   
LSTM can be easily connected to CNNs. Fully connected layers of a CNN can be directly fed into the input of LSTM blocks. To better represent the fully connected features, multiple LSTM kernels can act as a layer to represent temporal features. As shown in Figure \ref{lstm_array}, the CNN can extract the image features as a 1-D vector. The first frame of an image sequence at time $t_{1}$ is sent to the LSTM layer at $t_{1}$. The  LSTM layer will produce output feature $h_{1}$ for the first frame, then at time $t_{2}$, a new frame is sent to the LSTM layer and the new output feature is produced based on $x_{2}$ and $h_{1}$, so on so forth. Here we use $h_{i} (i=1...n)$ to represent the i$t$h LSTM feature; in Figure \ref{lstm_array} $n$ = 24. In different tasks, either only the last LSTM feature $h_{n}$ or the whole LSTM features ${\{h_{1}, h_{2}, ...h_{n}\}}$ are used for final prediction. In our case, we believe that all the frames can contribute to the AU detection. \color{black}Therefore, we use all the LSTM features; in our experiments, the number of frames is 24. \color{black}

LSTM can effectively fuse the temporal information in a sequence. Similar to the convolutional layers, more than one LSTM layers can be stacked to form an LSTM Net in order to achieve a deeper understanding of the temporal relationships. As shown in Figure \ref{lstm_array}, the LSTM Net has 2 LSTM layers  stacked for AU detection. To see if LSTM is useful in AU detection, we have conducted experiments to compare LSTM-based temporal fusion versus static image AU prediction. In order to find the best structure of LSTM, we also compared different depth of LSTM layers in Section 5.

\section{Experimental Evaluation}

\subsection{Datasets and Metrics}

\textbf{Dataset.}
AU datasets are harder to obtain compared to other tasks such as image classification. The reason is that there are multiple AUs on one face which requires much more manual labeling work. Here we give a brief review of the AU datasets referred by and compared in this paper.

(1) DISFA: 26 human subjects are involved in the DISFA dataset. The subjects are asked to watch videos while spontaneous facial expressions are obtained. The AUs are labeled with intensities from 0 to 5. We can obtain more than 100,000 AU-labeled images from the videos, but there are much more inactive images than the active ones. The diversity of people also makes it hard to train a robust model. 

(2) BP4D:  There are 23 female and 18 male young adults involved in the BP4D dataset. Both 2D and 3D videos are captured while the subjects show different facial expressions. Each subject participates in 8 sessions of experiments, so there are 328 videos captured in total. AUs are labeled by watching the videos. The number of valid AU frames in each video varies from several hundred to thousands. There are around 140,000 images with AU labels that we could use. 

To train a deep learning model, we need a larger number of image samples, and the diversity of the samples is also important. Following a common experimental setting in the AU detection community, we choose BP4D to train our model and conduct a 3-fold cross validation. We first split the dataset into 3 folds based on subject IDs. Each time two folds are used for training and the third fold for testing. For the DISFA dataset, we use the trained model from BP4D to directly extract the last fully connected layer feature with a length of 2048 to represent the images in DISFA. We run the same cross-validation evaluation experiments as the ones we performed with BP4D based on the extracted features using BP4D.

\textbf{Metrics.} 
One part of our task is to detect if the AUs are active or not, which is a multi-label binary classification problem. For a binary classification task especially when samples are not balanced, F1 score can better describe the performance of the algorithm \cite{p30,p10}. 
%
In our evaluation, we compute F1 scores for 12 AUs in BP4D and 8 AUs in DISFA. F1 scores can be compared directly as an indicator of the performance of different algorithms on each AU. The overall performance of the algorithm is described by the average F1 score.

\subsection{Adaptive Learning vs. Conventional CNN}

We proposed our ROI Nets for the adaptive region learning in Section 3. Compared to the conventional CNNs which share the same set of convolutional filters for the whole feature map, we hypothesize that by learning ROIs separately, a better understanding of AUs can be achieved. To validate this hypothesis, we train 2 neural networks on the BP4D dataset: a fine-tuned VGG model - FVGG, and the ROI Nets (on top of the basic VGG model). 12 AUs are used together, so the loss function is based on the predicted results for the 12 AUs. To prevent extreme loss explode which will stop the training, we added offsets to the loss function as shown by Equation \ref{loss}, where $l$ is the label and $p$ is the generated probability for an AU.

\begin{equation}
\label{loss}
Loss=-\Sigma(l\cdot\log(\frac{p+0.05}{1.05})+(1-l)\cdot\log(\frac{1.05-p}{1.05}))
\end{equation}\

The two models are both based on static images. During each iteration, we randomly select 50 images as a batch to compute the training loss. SGD is employed for back propagation. The VGG net pretrained parameters are used for initializing the model, and the parameters of the first 8 convolutional layers are not updated during training. This makes the set of parameters smaller, which helps the training algorithm converge. We use the proposed structure (VGG Net + ROI Nets) in Section 3 to train the adaptive region learning mode - which we still call ROI Nets. The new designed regional convolutional filters are initialized following a gaussian distribution. \color {black} For the conventional fine-tuned VGG (FVGG) net, only the last prediction layer of the basic VGG model is replaced with a fully-connected layer with 12 kernels. \color {black} We use sigmoid activation functions for the 12 AU probability generators. The two deep models both start with the same learning rate 0.001 which is decreased when the loss is stable. Momentum for both models is set to 0.9.

\begin{figure}[thpb]
      \centering
      \includegraphics[scale=.5]{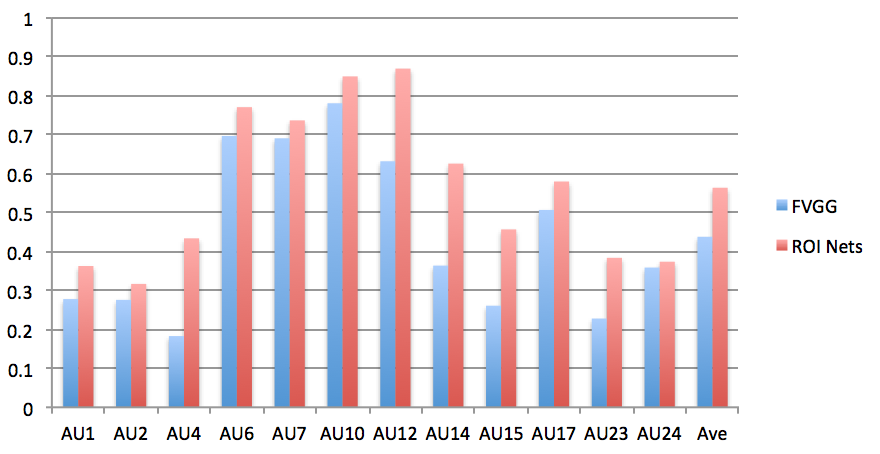}
      \caption{Comparison of FVGG and ROI-Nets in AU detection on BP4D}
      \label{vggvsrc}
   \end{figure}

The final models of both ROI Nets and FVGG are obtained after training the deep net 20,000 times. We then compare the F1 scores for each AU. The results are shown in Figure \ref{vggvsrc}. We can see that region learning with ROI Nets yields significant improvement\color {black}, on average by 12.4\%\color {black}. 

\begin{figure}[thpb]
      \centering
      \includegraphics[scale=.5]{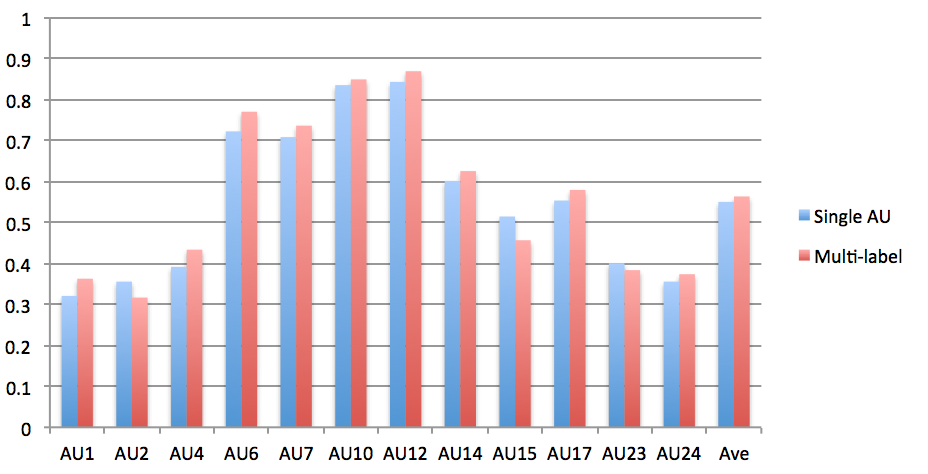}
      \caption{Comparison of single and multi-label learning on BP4D}
      \label{single-multilabel}
   \end{figure}

  \begin{figure*}[thpb]
      \centering
      \includegraphics[scale=.65]{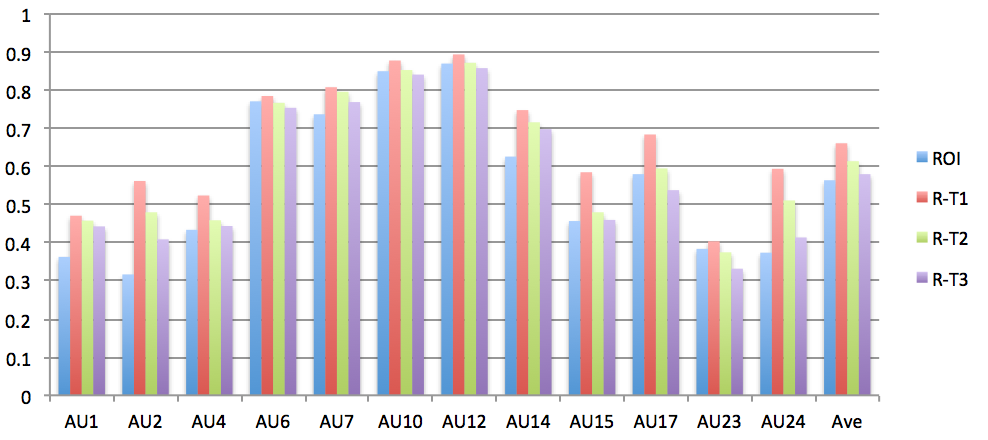}
      \caption{Comparison of static image and temporal fusion in AU detection on BP4D}
      \label{lstms_rst}
   \end{figure*}

\subsection{Single vs. Multi-label AU Detection}

In our proposed ROI Nets, the regions are determined based on the positions where the AUs take place. Since each AU has corresponding regions, we may use only the local learned features to represent the AU for detection. This single AU detection approach differs from the approach we use for the adaptive region learning evaluation (Figure \ref{fig_cnn_all}) where we concatenate all the AUs features as one fused feature. 
Our hypothesis is, by concatenating multiple AU features, we may obtain valuable global information as a supplement for individual AU detection or to provide more correlations. However, it's also possible that it brings some noise to the ``purity" of an AU feature. To validate our hypothesis, we conduct an experiment to compare single AU detection and multi-label AU detection. In multi-label AU detection, one image is labeled with multiple AUs. In this case, we cannot guarantee that we are able to provide the same number of positive and negative samples for all AUs. But for single AU detection, since the training for each AU is performed separately, we can prepare the training data for each AU in a way that the training data is always balanced during training. The AU detection results for single vs multiple AU detection is shown in Figure \ref{single-multilabel}.

By comparison, we can clearly see that even with equal positive and negative sample distribution, the multi-label AU detection slightly outperforms the single AU detection approach in most AUs\color{black}, on average by 1.3\%\color{black}. That implies that the global information does have an important impact on the fusion learning. We have some more interesting findings if we look into the different AU detection results. For the under-represented AUs (where the AU shows up less frequently in the dataset), such as AU2, AU15, AU23, the balancing of training samples (as in the single AU detection) can boost the performance more significantly. Whereas for some highly related AUs such as AU6 and AU12, both for happy, the multi-label learning has a higher chance to learn this correlation and improve the AU detection for these two AUs.

\subsection{Temporal vs. Static}

\begin{table*}
\caption{F1 score on BP4D dataset (ROI: ROI Nets; R-Ti: ROI Nets + i-layer LSTM Net )}
\label{tab_bp4d}
\begin{center}
\rowcolors{2}{gray!25}{white}
\begin{tabular}{|c|c|c|c|c|c|c|c|c|c|c|}
\hline
AU& LSVM& JPML\cite{p13}& DRML\cite{p22}&CPM\cite{p18}&CNN+LSTM\cite{p31}&FVGG &ROI &R-T1&R-T2&FERA\cite{jaiswal2016deep}\\
\hline
1& 23.2& 32.6& 36.4&43.4 &31.4 &27.8 &36.2 &{\bf47.1} & 45.8& 28\\
2& 22.8& 25.6&  41.8& 40.7& 31.1 & 27.6&31.6  &{\bf56.2} & 48.0&28\\
4& 23.1& 37.4& 43.0& 43.4 &{\bf71.4} &18.3 &43.4 & 52.4 &45.9&34\\
6& 27.2& 42.3& 55.0&59.2  &63.3 &69.7&77.1 &{\bf 78.5}&76.7&70\\
7& 47.1& 50.5& 67.0& 61.3 &77.1 &69.1& 73.7&{\bf 80.8}&79.6&78 \\
10&77.2&72.2&66.3&62.1 &45.0 &78.1&85.0&{\bf 87.8}&85.3&81\\
12&63.7&74.1&65.8&68.5 &82.6 &63.2&87.0&{\bf89.4 }&87.2&78\\
14&64.3&65.7&54.1&52.5 &72.9 &36.4&62.6&{\bf 74.8}&71.6&75\\
15&18.4&38.1&36.7 &34.0 &33.2&26.1&45.7 &{\bf58.5} &48.0&20\\
17&33.0&40.0&48.0& 54.3&53.9 &50.7&58.0&{\bf 68.4} &59.5&36\\
23&19.4&30.4&31.7& 39.5& 38.6&22.8&38.3&{\bf 40.4} &37.5&41\\
24&20.7&42.3&30.0& 37.8& 37.0&35.9&37.4&{\bf 59.4} &51.1&-\\
Avg&35.3&45.9&48.3& 50.0&53.2 &43.8&56.4&{\bf66.1} &61.4& 51.7\\
\hline
\end{tabular}
\end{center}
\end{table*}

A facial action always has a temporal component, hence knowing the previous state of a facial expression can definitely improve the AU detection. We proposed the LSTM layer for fusing the temporal information with static image features. From our previous evaluations, the best performance was obtained for static images with the ROI Nets. In this experiment, we use the ROI model as a baseline to compare with region cropping recurrent temporal model (noted as R-T in figures and tables). 
Here, the LSTM layers are used for fusing the static image features. 512 LSTM kernels are employed to construct each LSTM layer. We then utilize 24 frames as a sequence to represent the video. In our data preparation, we follow the same framework as the one we used to train the static image learning models. The only difference is that to construct the image sequence, we randomly find other 23 images prior to the selected image from the same subject. This will create more non-repeatable training data. Afterward, the sequence is fed into the training model. To find the best LSTM structure, we tried 1 (in R-T1), 2 (in R-T2) and 3 (in R-T3) stacked LSTM layers for AU detection, as demonstrated in Figure \ref{lstm_array}. The AU detection results are shown in Figure \ref{lstms_rst}.  

From the results shown in Figure \ref{lstms_rst}, we can clearly observe the improvement in AU detection due to applying the LSTM layers. The average F1 score is also improved by \color {black} 9.7\% using R-T1 over ROI Nets\color{black}. Another conclusion we can make here is that with more LSTM layers, the performance decreases, as the ROI features are sufficient to represent the AU images and one LSTM layer is enough to reveal the temporal corrections.

\subsection{Performanc Comparison}

\begin{table}
\caption{F1 score on DISFA dataset}
\label{tab_disfa}
\centering
\tabcolsep=0.11cm
\rowcolors{2}{gray!25}{white}
\begin{tabular}{|c|c|c|c|c|c|c|}
\hline
AU&LSVM&APL\cite{p22}&DRML\cite{p22}&FVGG&ROI&R-T1\\
\hline
1&10.8&11.4&17.3&32.5&41.5&\bf{42.6}\\
2&10.0&12.0&17.7&24.3&26.4&\bf{27.2}\\
4&21.8&30.1&37.4&61.0&\bf{66.4}&65.5\\
6&15.7&12.4&29.0&34.2&50.7&\bf{55.5}\\
9&11.5&10.1&10.7&1.67&8.5&\bf{22.8}\\
12&70.4&65.9&37.7&72.1&\bf{89.3}&82.9\\
25&12.0&21.4&38.5&87.3&\bf{88.9}&88.3\\
26&22.1&26.9&20.1&7.1&15.6&\bf{25.9}\\
Avg&21.8&23.8&26.7&40.2&48.5&\bf{51.3}\\
\hline
\end{tabular}
\end{table}

By observing the results of our previous experiments, we can clearly see that the ROI Nets can learn more powerful local AU features that would result in better AU detection compared to conventional CNNs. The performance was similar to single AU detection and multi-label AU detection, but the multi-label detection approach shows slightly better overall performance due to the strong correlation among AUs and richer global information. In the static/temporal exploration experiment, we witnessed that the LSTM Net with one LSTM layer boosts the AU detection accuracy by a 9.7\% average F1 improvement, which implies that the temporal context information plays a very important role in detecting facial actions. 

To compare our approaches with another state of the art methods, we have collected the F1 measures of the most popular methods in same 3-fold settings based on BP4D in Table \ref{tab_bp4d}. The approaches includes a traditional SVM-based method, a 2-D landmark feature based approach, JPML \cite{p13}, the Confidence Preserving Machine (CPM) \cite{p18}, a block-based region learning static CNN, DRML \cite{p22}, and a recurrent net fusing LSTM with simple CNN, CNN+LSTM \cite{p31}. For our proposed approaches, we first use the FVGG as the baseline approach. Then we show the results of adaptive ROI Nets based on static images. Finally, we test our ROI Nets + our LSTM based recurrent approach with one and two LSTM layers (RC+T1, RC+T2). All the results can be seen in Table \ref{tab_bp4d}. On average, our best model R-T1 achieves a 12.9\% improvement compared to the state of the art approach. \color{black}Across the 12 AUs, our R-T1 model outperforms the best in the literature except for AU4, where CNN+LSTM performs the best. 

To further explore the capabilities of our proposed approach, we run the comparison on DISFA dataset as well. Not as popular as BP4D, a fewer state of the art approaches report their results on DISFA. We use the BP4D trained model to extract features from all the images in DISFA and conduct a 3-fold cross evaluation with the extracted features. For static image evaluation, we directly run a multi-label linear regression and for temporal evaluation, we use the structure that shows the best performance in the BP4D evaluation, that is, a one layer LSTM to train the DISFA temporal model. The results are shown in Table \ref{tab_disfa}. As we can see, our R-T1 model leads to a 25\% improvement over the state of the art model. 

From the results in Tables \ref{tab_bp4d} and \ref{tab_disfa}, our proposed approaches have the best performance in both static and sequence image based AU detection. 
In the static images based AU detection using deep learning, our ROI Nets outperforms the state of the art deep learning approach, DRML. Our proposed adaptive region cropping method shares the same idea of learning different sets of convolutional filters for different sub-regions, but our method has the following advantages that make it different from the state of the art: 

1) Our sub-region selection is adaptive.  
DRML used a straightforward image dividing strategy. Assuming the facial images are aligned, each image is equally divided into 8x8=64 sub-regions. This framework in easy to implement, but we need to make sure that the face images are actually aligned in the first place. In order to assure this precondition, all the faces need to be transformed to a neutral shape. This may cause information loss since the faces of different individuals may have different shapes or sizes. In addition, if the original faces are not in a frontal pose, we may also lose some appearance features after changing the pose. On the contrary, we select the regions of interest adaptively. Our approach works based on the detected landmarks and the positions of facial action muscles, which are biologically meaningful. Also note that our approach is robust to landmark position errors. This is because the feature maps in our network go through several pooling layers. Imagine that the position detection error in the original image of size 224$\times$224 is 10 pixel. With the pooling layer for cropping the feature map being of size 14$\times$14, the error turns to be less than 1 pixel. This significantly improves our proposed adaptive region cropping net.

2) A very deep pretrained network (VGG) is used as the base. 
DRML creates a shallow convolutional network for the region based AU detection. Instead of training everything from scratch, we choose to borrow parameters from an existing very deep CNN model. The main advantage of this approach is that the pretrained model has been trained with millions of images. Although the tasks are different, the parameters are transferable. With the pretrained model as the starting point of our AU detection training, we can achieve a more powerful model than by training a shallow neural network. 

In sequential image based AU detection, Chu et al. \cite{p32} designed a network by combining both CNN and LSTM. To obtain the spatiotemporal fusion features, the last layer features of the CNN and LSTM nets are concatenated. Different from their use of AlexNet for static image feature extraction, we have proposed the adaptive region cropping convolutional net. We use LSTM to fuse the temporal deep features as well, but we have also compared different layers of LSTM and noticed that one layer LSTM shows the best performance.

\section{Conclusion}

In this paper, we looked into three essential problems, the region adaption learning, temporal fusion and single/multi-label AU learning, in AU detection and proposed a novel approach to address these problems. We first proposed an adaptive region of interest cropping net, which compared to conventional CNN, proves to be able to learn separate filters for different regions and can improve the accuracy of AU detection. We then analyzed the proposed model by training it in a multi-label AU detection manner and showed that the new model can outperform a single AU detection model. We finally explored the LSTM-based temporal fusion approach, which boosted the AU detection performance significantly compared to static image-based approaches. We also tried to find an optimal structure of LSTM layers to connect with the proposed ROI nets to achieve the best results for AU detection. The proposed approach is evaluated on two popular AU detection datasets, BP4D and DISFA, outperforming the state of the art significantly, with an average improvement of around 13\% and 25\% on BP4D and DISFA respectively. Our future work will be focused on building a dataset-independent AU detection model and applying it to facial action detection in real world applications. 
\section{Acknowledgement}
This work is supported by the National Science Foundation through Award EFRI -1137172, and VentureWell (formerly NCIIA) through Award 10087-12.
{\small
\bibliographystyle{ieee}
\bibliography{egbib}
}

\end{document}